# 图像复原中自注意力和卷积的动态关联学习

江奎[1]，贾雪梅[§2]，黄文心[*3]，王文兵[4]，王正[2]，江俊君[1]

1. 哈尔滨工业大学,哈尔滨 150000； 2. 武汉大学,武汉 430072 3. 湖北大学,武汉 430062 4. 杭州灵伴科技有限公司，杭州 310000

**摘　要：目的** 卷积神经网络(Convolutional Neural Network, CNN)和自注意力（Self-attention, SA）在多媒体应用领域已经取得了巨大的成功。然而，鲜有研究人员能够在图像修复任务中有效地协调这两种架构。针对这两种架构各自的优缺点，本文提出了一种关联学习的方式以综合利用两种方法的优点并抑制各自的不足，实现高质高效的图像修复。**方法** 本文结合CNN和SA两种架构的优势，尤其是在特定的局部上下文和全局结构表示中充分利用CNN的局部感知和平移不变性，以及SA的全局聚合能力。此外，图像的降质分布揭示了图像空间中退化的位置和程度。受此启发，本文在背景修复中引入退化先验，并据此提出一种动态关联学习的图像修复方法。核心是一个新的多输入注意力模块，将降质扰动的消除和背景修复关联起来。通过结合深度可分离卷积，利用CNN和SA两种架构的优势实现高效率和高质量图像修复。**结果** 在Test1200数据集中进行了消融实验以验证算法各个部分的有效性，实验结果证明CNN和SA的融合可以有效提升模型的表达能力；同时，降质扰动的消除和背景修复关联学习可以有效提升整体的修复效果。本文提出的方法在三个图像修复任务的合成和真实数据上与最新的10余种方法进行了比较，提出的方法取得了显著的提升。在图像去雨任务上，本文提出的ELF方法在合成数据集Test1200上，相比于MPRNet，PSNR值提高了0.9 dB；在水下图像增强任务上，ELF在R90数据集上超过Ucolor方法4.15dB；在低照度图像增强任务上，相对于LLFlow算法，ELF获得了1.09 dB的提升。**结论** 本文提出的方法兼具高效果和性能，在常见的图像去雨、低照度图像增强和水下图像修复等任务上优于代表性的方法。

# Dynamic Association Learning of Self-Attention and Convolution in Image Restoration

Kui Jiang[1]，Xuemei Jia[2]，Wenxin Huang[3]，Wenbin Wang[4]，Zheng Wang[2]，Junjun Jiang[1]

1. Harbin Institute of Technology, Harbin 150000； 2. Wuhan University, Wuhan 430072

3. Hubei University, Wuhan 430062 4. Hangzhou Lingban Technology Ltd., Hangzhou 310000

**Abstract: Objective** Convolutional Neural Network (CNN) and Self attention (SA) have achieved great success in the field of multimedia applications for dynamic association learning of self-attention and convolution in image restoration. However, due to

*通信作者：黄文心 wenxinhuang_wh@163.com

§共同第一作者：贾雪梅
**基金项目**：国家自然科学基金(6230010538, 61971165)；湖北省重点研发基金(2021YFC3320301)；中国人工智能学会-华为MindSpore 开放基金(2022BAA033).

Supported by: National Natural Science Foundation of China (6230010538, 61971165), Hubei Key R\&D Project (2021YFC3320301), CAAI-Huawei MindSpore Open Fund (2022BAA033).

intrinsic characteristics of local connectivity and translation equivariance, CNNs have at least two shortcomings: 1) limited receptive field; 2) static weight of sliding window at inference, unable to cope with the content diversity. The former thus prevents the network from capturing the long-range pixel dependencies while the latter sacrifices the adaptability to the input contents. As a result, it is far from meeting the requirement in modeling the global rain distribution, and generates results with obvious rain residue. Meanwhile, due to the global calculation of SA, its computation complexity grows quadratically with the spatial resolution, making it infeasible to apply to high-resolution images. In view of the advantages and disadvantages of these two architectures, this paper proposes an association learning method to comprehensively utilize the advantages of the two methods and suppress their respective shortcomings, so as to achieve high-quality and efficient inpainting. **Method** This article combines the advantages of both CNN and SA architectures, especially by fully utilizing CNN's local perception and translation invariance in specific local context and global structural representations, as well as SA's global aggregation ability. We take inspiration from the observation that rain distribution reflects the degradation location and degree, in addition to the rain distribution prediction. Therefore, we propose to refine background textures with the predicted degradation prior in an association learning manner. As a result, we accomplish the image deraining by associating rain streak removal and background recovery, where an image deraining network (IDN) and a background recovery network (BRN) are specifically designed for these two subtasks. The key part of association learning is a novel multi-input attention module (MAM). It generates the degradation prior and produces the degradation mask according to the predicted rainy distribution. Benefited from the global correlation calculation of SA, MAM can extract the informative complementary components from the rainy input (query) with the degradation mask (key), and then help accurate texture restoration. Meanwhile, SA tends to aggregate feature maps with self-attention importance, but convolution diversifies them to focus on the local textures. Unlike Restormer equipped with pure Transformer blocks, we promote the design paradigm in a parallel manner of SA and CNNs, and propose a hybrid fusion network. It involves one residual Transformer branch (RTB) and one encoder-decoder branch (EDB). The former takes a few learnable tokens (feature channels) as input and stacks multi-head attention and feed-forward networks to encode global features of the image. The latter, conversely, leverages the multi-scale encoder-decoder to represent contexture knowledge. We propose a light-weight hybrid fusion block (HFB) to aggregate the outcomes of RTB and EDB to yield a final solution to the subtask. In this way, we construct our final model as a two-stage Transformer-based method, namely ELF, for single image deraining. **Result** A ablation experiment was conducted on the Test1200 dataset to validate the effectiveness of various parts of the algorithm. The experimental results showed that the fusion of CNN and SA can effectively improve the model's expression ability; At the same time, the elimination of degraded disturbances and background repair association learning can effectively improve the overall repair effect. The method proposed in this paper is compared with more than 10 new methods on the synthesis and real data of three inpainting tasks, and the proposed method has achieved significant improvement. In the task of image rain removal, the ELF method proposed in this paper improves the PSNR value by 0.9 dB compared to MPRNet on the synthesized dataset Test1200; In the underwater enhancement task, the ELF exceeds the Ucolor by 4.15dB on R90 dataset; In the low illumination image enhancement task, ELF achieves a 1.09 dB improvement compared to the LLFlow algorithm. **Conclusion** We rethink the image deraining as a composite task of rain streak removal, textures recovery and their association learning, and propose an ELF model for image deraining. Accordingly, a two-stage architecture and an associated learning module (ALM) are adopted in ELF to account for twin goals of rain streak removal and texture reconstruction while facilitating the learning capability. Meanwhile, the joint optimization promotes the compatibility while maintaining the model compactness. Extensive results on image deraining and joint detection task demonstrate the superiority of our ELF model over the state-of-the-arts. The method proposed in this paper possesses efficiency and effectiveness, and is superior to representative methods in common tasks such as image rain removal, low-light image enhancement and underwater enhancement.

# 0 引 言

复杂的成像条件，如雨雾、低光、水下散射等会对图像质量产生不利影响，并显著降低基于人工智能应用技术的性能，如图像理解（Liang 等，2022；Wang 等 2022）、目标检测（Zhong 等，2021）和目标识别（Xie 等，2022）。因此，急需研究有效的图像修复方案，消除成像过程中的降质扰动，提升图像的可辨识度和可读性，输出高质量的修复结果。

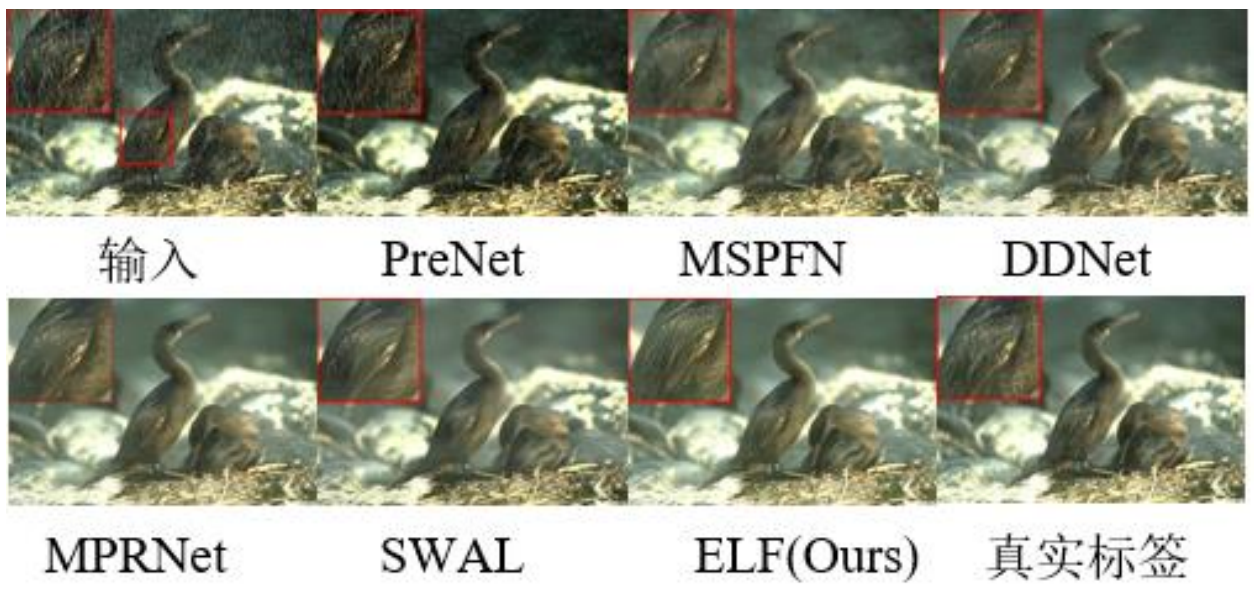

图1 各种去雨方法的结果比较

Fig.1 Comparison of the results of various deraining methods.

在过去的十年中，图像修复（Ma 等，2018；Chen 等，2021；Wang 等，2020；Yang 等，2022）获得了前所未有的发展。在深度神经网络之前，基于模型的图像修复方法（Garg 等，2005）更多地依赖于图像内容的统计分析，并在降质或者背景上引入人为设定的先验知识（例如稀疏性和非局部均值滤波）。尽管如此，这些方法在复杂多变的降质环境中稳定性较差（Bossu 等，2011；Chen 等，2013；Zhong 等，2022）。

与传统基于模型的方法相比，卷积神经网络（Convolutional Neural Network，CNN）能够从大规模的数据中学习到广义统计知识，无疑是更好的选择。为了进一步提高图像修复的效果，现有网络设计了各种复杂的结构和训练方式（Jiang 等，2020；Yang 等，2021；Yu 等，2019）。然而，由于局部感知和平移同变性的固有特征，CNN 至少有两个缺点：1）感受野有限；2）滑动窗口在推理时的静态权重无法应对内容的多样性。具体来讲，前者使网络无法捕捉到长距离的像素依赖性，而后者则牺牲了对输入内容的适应性。因此，它远远不能满足表征全局降质分布的需求。以图像去雨为例，上述基于 CNN 的方法输出结果会有明显的雨水残留((Ren 等，2019）和 DRDNet（Jiang 等，2020))或细节损失（MPRNet（Zamir 等，2021）和 SWAL（Huang 等，2021))（如图 1 中的去雨结果所示）。

给定一个像素，自注意力（Self-Attention，SA）会通过其他位置的加权去获得当前位置的全局响应。在各种自然语言和计算机视觉任务的深度网络中都进行了相关的研究（Vaswani 等，2017；Wang 等，2018；Zhang 等，2019）。得益于全局处理的优势，SA 在消除图像扰动方面取得了比 CNNs 更加显著的性能提升（Chen 等，2021；Liang 等，2021；Wang 等，2022）。然而，由于 SA 的计算是全局的，其计算复杂度随空间分辨率呈二次方增长，因此无法应用于高分辨率图像。最近，SA 也被应用于图像修复任务，如图像去雨，去雾，超分等。Restormer（Zamir 等，2022）提出了一种多头转置注意（Multi-Dconv head Transposed Attention，MDTA）模块来建模全局关联，并取得了令人印象深刻的图像修复效果。尽管 MDTA 是在特征维度上而不是在空间维度上应用 SA，具有线性的复杂度，但 Restormer（Zamir 等，2022）还是需要更多的计算资源才能获得更好的恢复性能。因其具有 563.96 GFlops 和 2610 万个参数，使用一个 TITAN X GPU 对 512×512 像素的图像进行去雨需要 0.568 秒，这对于许多资源有限的实际应用来讲，所需的算力或内存都是昂贵的。

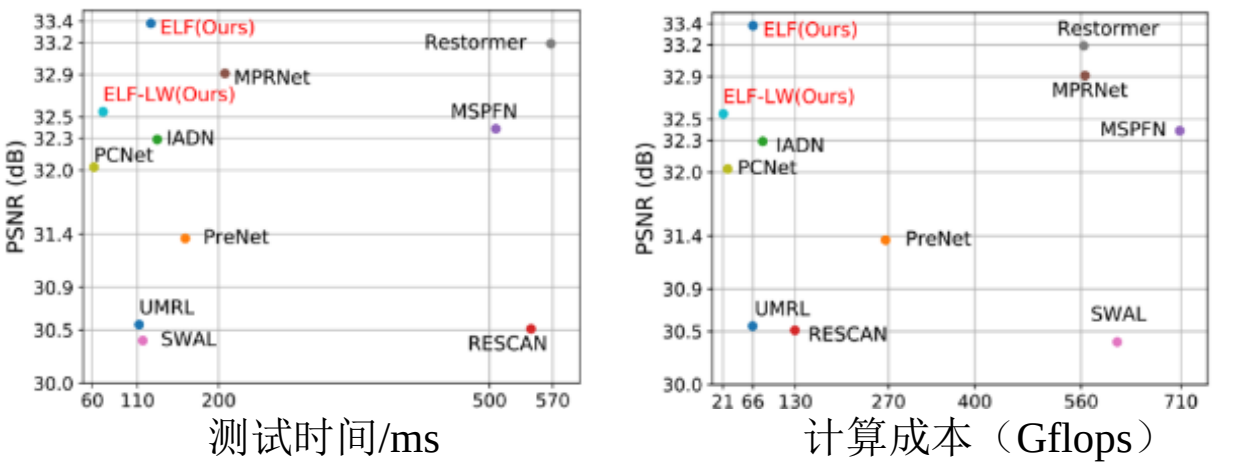

图 2 在 TEST1200 数据集上比较主流图像去雨方法的效果与性能

Fig.2 Comparison of mainstream deraining methods in terms of efficiency vs. performance on the TEST1200 dataset.

除效率低之外，Restormer 至少还有两个缺点。1)将图像修复看作是基于扰动和背景图像的简单叠加，这是有争议的。因为降质扰动层和背景层是交织重叠的，其中降质影响了图像的内容，包括细节、颜色和对比度。2）构建一个完全基于 Transformer 的框架是次优的。因为 SA 擅长聚合全局特征图，但缺乏 CNNs 在学习局部上下文关系方面的能力。这自然引出了两个问题：（1）如何将降质扰动去除与背景修复联系起来？（2）如何将 SA 和 CNNs 有效地结合起来实现高精度和高效率的图像修复？

为了解决第一个问题，本文从降质分布揭示退化位置和程度的观测中得到启示，降质分布反映了图像退化的位置的强度。因此，本文提出以关联学

习的方式，利用预测的退化来优化背景纹理重建，将扰动去除与背景重构相结合，分别设计了图像雨纹移除网络（Image Deraining Network，IDN）和背景重构网络（Background Recovery Network，BRN）来完成图像修复。关联学习的关键部分是一种新的多输入注意模块 MAM。它对输入降质图像中得到的退化分布进行量化表征，生成退化掩码。得益于SA 的全局相关性计算，MAM 可以根据退化掩码从降质输入中提取背景信息，进而有助于网络准确的恢复纹理。

处理第二个问题的一个直观想法是利用这两种架构的优势构建一个统一的模型。Park 等（2022）已经证明 SA 和标准卷积网络有着相反且互补的特性。具体来说，SA 倾向于聚合具有自注意力中重要的特征图，但卷积使其多样化，以专注于局部纹理。与 Restormer 中设置的 Transformer 不同，本文以并行的方式处理 SA 和 CNN，并提出了一种交叉融合网络。它包括一个残差 Transformer 分支（Residual Transformer Branch，RTB）和一个编码器-解码器（Encoder-Decoder Branch，EDB）。前者通过多头注意力和前馈网络来编码图像的全局特征。相反，后者利用多尺度编码器-解码器来表示上下文知识。并且本文设计了一种轻量级交叉融合块（Hybrid Fusion Block，HFB）来聚合 RTB 和 EDB 的结果，最终用以处理对应的学习任务。通过这种方式，最终构建一种基于 Transformer 的两阶段模型，即ELF。在图像去雨任务上，其平均性能优于基于 CNN 的 SOTA（MPRNet（Zamir 等，2021））0.25dB，并且节省了 88.3%和 57.9%的计算成本和参数。

本文的主要贡献如下：

（1）本文是首次考虑到 Transformer 和 CNN 在图像修复任务中的高效性和兼容性，并将 SA 和CNN 的优势整合到一个基于关联学习的网络中，用于扰动消除和背景重构。这是一个针对图像修复任务的局部-整体多层次结构的高效实现。

（2）本文设计了一种新的多输入注意力模块（MAM），将扰动去除和背景重构任务巧妙地关联起来。它显著减轻了网络学习负担，同时促进了背景纹理恢复。

（3）在图像去雨、水下图像增强、低光照增强和检测任务上的综合实验论证了本文提出的 ELF 方法的有效性和效率。以图像去雨任务为例，ELF 平均比 MPRNet（Zamir 等，2021）高出 0.25dB，而后者的计算成本为 8.5 倍，参数为 2.4 倍。

# 1 相关工作

在过去几年中，图像去雨的相关工作在架构创新和训练方法方面都取得了重大进展。接下来，本节将简要介绍一些典型的且与本文研究相关的图像去雨、图像恢复和视觉 Transformer 模型。

## 1.1 单图像去雨

传统的去雨方法（Kang 等，2011；Luo 等，2015）采用图像处理技术和手工制作的先验来解决去雨问题。然而，当预定义的模型不成立时，这些方法会产生较差的结果。最近，出现的基于深度学习的去雨方法（Li 等，2017；Zhang 等，2017；Jiang 等，2021）都表现出令人印象深刻的性能。早期基于深度学习的去雨方法（Fu 等，2017；Zhang 等，2018）应用卷积神经网络（CNN）直接减少从输入到输出的映射范围，以此产生无雨结果。为了更好地表示雨水分布，研究人员考虑了雨水特征，如雨密度（Zhang 等，2018）、大小和遮蔽效应（Li 等，2017；Li 等，2019），并使用递归神经网络通过多个阶段（Li 等，2018）或非局部网络（Wang 等，2020）来利用长距离空间相关性更好地去除雨纹（Li 等，2018）。在此基础上，SA 利用其强大的全局相关学习消除了雨水退化，取得了优秀的效果。虽然采用精简表示和基于全局不重叠窗口的 SA（Wang 等，2022；Ji 等，2021）来提升全局 SA 以减轻计算负担，但这些模型仍然会迅速占用计算资源。除了效率低之外，这些方法（Zamir 等，2022；Ji 等，2021）仅将去雨任务视为雨水扰动的消除，忽略了退化带来地背景细节缺失和对比度偏差。

## 1.2 图像恢复

从低质量图像中恢复高质量图像的任务被统称为图像恢复任务，如水下图像增强、低光照图像增强、图像去雾等，拥有与图像去雨类似的降质因素。接下来，本小节简要介绍一些典型的水下图像增强和低光照图像增强方法。

### 1.2.1 水下图像增强

早期的水下图像增强方法通过动态像素范围拉伸（Iqbal 等，2010），像素分布调整（Ghani 等，2015）和图像融合（Ancuti 等，2012）等方法来调节像素值以实现增强，但这些方法难以应对多样的水下场景。随着深度学习的发展，一些基于深度学习的水下图像增强方法逐渐被提出。其中基于生成对抗网络的方法成了主流，如 UCycleGAN（Li 等，2018）采用弱监督的方式将 CycleGAN（Zhu 等，

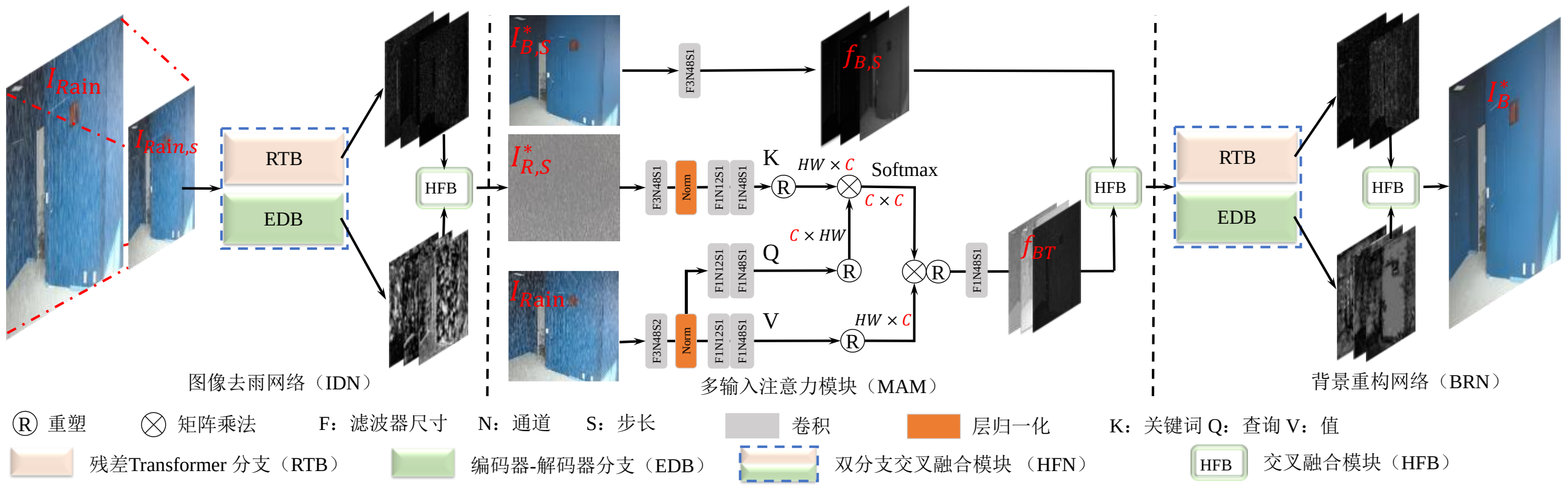

图 3 本文提出的图像修复方法 ELF 的网络结构（以图像去雨任务为例）

Fig.3 The architecture of our proposed ELF image restoration method (taking image deraining as an example).

2017）的网络结构应用到此任务中，Guo 等人（2019）提出一个多尺度密集生成对抗网络，都取得了不错的效果。但他们都只是简单应用了基于生成对抗网络的结构，并没有考虑复杂的退化关系，生成的结果有明显的雨水残留，而且会引入对比度失真。

1.2.2 低光照图像增强

早期的低光照图像增强方法多基于像素灰度值统计分析，如直方图均衡化（Cheng 等，2004；Pisano 等，1998）等。但这些方法只利用了灰度分布，并没有考虑空间信息，增强后的图像可能会过曝光或欠曝光，与真实图像不一致。相比之下，基于视网膜层理论（Retinal Cortex Theory）的方法（Jobson 等，1997）将输入的低光照图像分解为光照和反射率两部分，通过增强光照部分来增强图像。但这些方法通常缺乏足够的适应性，难以获得稳定的光照分布，且易缺失细节纹理信息。通过学习低光图像到正常光图像的映射，基于深度学习的方法取得了综合的最优效果。例如，Zero-DCE（Guo 等，2020）通过逐步推导构造出了一种轻量的像素级别的曲线估计网络，来学习像素级高阶曲线参数映射，同时提出无参考损失函数对输出图像的质量进行间接的评估。EnlightenGAN（Jiang 等，2021）提出了一种高效无监督的生成对抗网络，并对全局-局部鉴别器结构，自正规化感知损失融合和注意机制进行了测试，实现了很好的低光照图像增强效果和通用性。LLFlow（Wang 等，2022）提出以低光图像/特征为条件，学习将正常曝光图像的分布映射到高斯分布中。然后，通过在训练中约束正常图像的光流结构实现图像增强。但是，单一映射的网络结构使得他们在应对复杂输入时可能产生伪影、色差等问题，且难以恢复精细的结构纹理。

### 1.3 视觉 Transformer

基于 Transformer 的模型首先应用在自然语言任务中的序列处理（Vaswani 等，2017）。由于 ViT（Dosovitskiy 等，2020）具有很强的长距离依存关系学习能力，因此将 Transformer 引入了计算机视觉领域，并将大量基于 Transformer 的方法应用于计算机视觉任务，例如图像识别（Dosovitskiy 等，2020；Ijaz 等，2022），分割（Wang 等，2021），目标检测（Carion 等，2020；Liu 等，2021）。对于给定的输入内容（Khan 等，2021），视觉 Transformer（Dosovitskiy 等，2020；Touvron 等，2021）将一幅图像分解为一组局部窗口序列，并学习它们之间的相互关系。例如，TTSR（Yang 等，2020）提出了一种自注意力模块，可以提供准确的纹理特征，用于将参考图像中的纹理信息传输到高分辨率图像进行重建。Chen 等人（2021）在 ImageNet 数据集上提出了一个预训练的图像处理 Transformer，并使用多头网络架构分别处理不同的任务。然而，SA 的直接应用未能充分利用 Transformer 的潜力，这是由于自注意力巨大的计算负载和不同深度（尺度）层之间的低效通信造成的。此外，很少有工作考虑到 Transformer 与 CNNs 之间的内在互补特性去构建一个有效统一的模型。自然地，这种设计限制了局部邻域内的上下文融合表达，这违背了使用自注意力而不是卷积的主要动机，因此不太适合图像恢复任务。相比之下，本文探索连接两者的桥梁，并为图像去噪任务构建了 Transformer 和 CNN 的交叉模型。

## 2 本文方法

本文的主要目标是利用 CNN 和 Transformer 构建高效、高精度的图像修复模型。理论上，自注意力将特征映射值与正向的重要权重进行平均，以学习全局表示，而 CNN 倾向于聚合局部相关信息。

直观上，将它们结合起来以充分利用局部和全局纹理是合理的。一些研究试图将这两种结构结合起来，形成一种用于浅层图像恢复的交叉框架，但是未能充分发挥其作用。

以图像去雨为例，与直接将 Transformer 块替换卷积的方法不同，本文考虑了这两种结构的高效性和兼容性，并构建了一个称为 ELF 的交叉框架，能够充分协调它们在图像修复任务上的优势。与现有的图像修复方法相比，所提出的 ELF 至少在两个关键的方面与它们不同。（a）设计概念的不同：与基于叠加模型的方法不同的是，ELF 将背景图像 $I_B$ 的最优近似值 $I_B^*$ 从雨天图像 $I_{Rain}$ 中预测出来，或从雨天图像中残差学习雨水信息 $I_R$ 并生成 $I_B^*$，ELF 将图像去雨任务转换为雨纹去除和背景重构的组合，并引入 Transformer 将这两部分与新设计的多输入注意力模块（MAM）联系起来。成分的不同：由于低频信号和高频信号是 SA 和卷积（Park 等，2022）中十分重要的信息，因此构建了一个用于特定特征表示和融合的双分支框架。具体来说，ELF 的主干是一个双分支交叉的融合网络，包括了一个残差 Transformer 分支（RTB）和一个编码器-解码器分支（EDB），分别学习全局结构（低频成分）的表征和局部纹理（高频成分）的表征。

图 3 概述了提出的 ELF 的框架，该框架包含图像去雨网络（IDN）、多输入注意力模块（MAM）和背景重构网络（BRN）。为提高效率，IDN 和 BRN 共享相同的双分支交叉融合网络，详见第 2.2 节。

## 2.1 网络流程及优化

给定一张雨天图片 $I_{Rain} \in R^{H \times W \times 3}$ 和一张干净版本的图片 $I_B \in B^{H \times W \times 3}$，其中 $H$ 和 $W$ 表示映射特征的空间高度和宽度。可以观察到，雨图样本 $I_{Rain,S} \in R$ 经过双线性插值重建的雨天图像 $I_{Rain,SR} \in R^{H \times W \times 3}$ 与原始雨天图片有着相似的统计分布，如图 4 所示。受到启发，本文在样本空间中去预测雨纹分布，以减轻学习和计算负担。

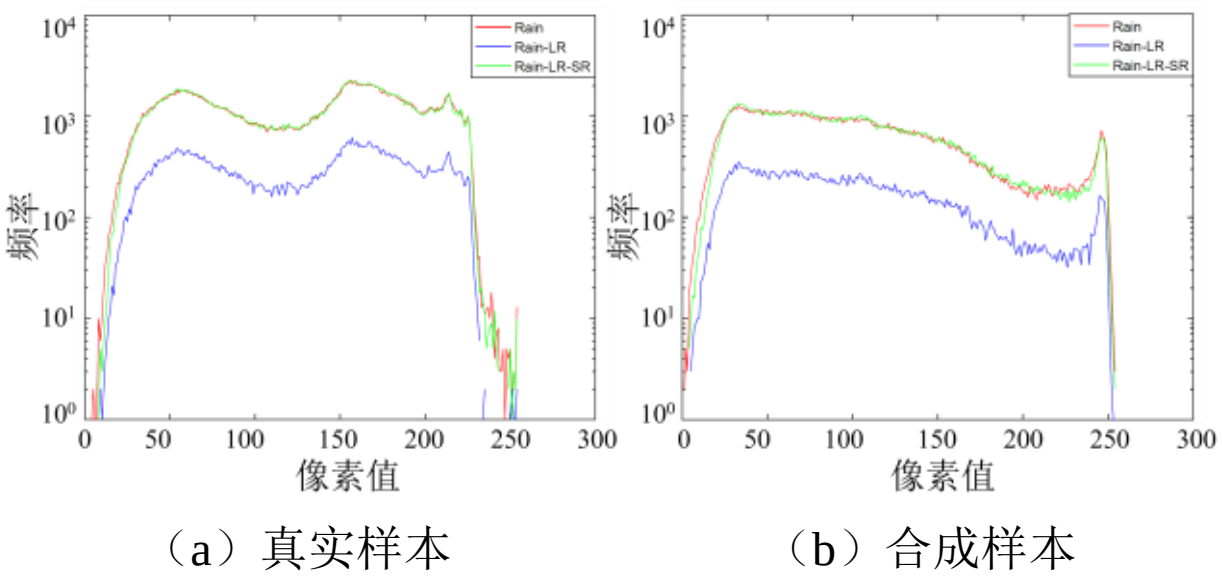

（a）真实样本　　　（b）合成样本

图 4 真实样本与合成样本的“Y”通道直方图拟合结果

Fig.4 Fitting results of “Y" channel histogram for Real and Synthetic samples

以这种方式，首先对 $I_{Rain}$ 和 $I_B$ 进行双线性操作，生成相应的子样本（$I_{Rain,S} \in R$ 和 $I_{B,S} \in B$）。如前所述，ELF 包含两个子网络（IDN 和 BRN），通过关联学习来完成图像去雨。因此，$I_{Rain,S}$ 被输入到 IDN 中，生成相应的雨水分布 $I^*_{R,S}$ 和去雨结果 $I^*_{B,S}$，

$$I^*_{R,S} = \vartheta_{IDN}(F_{BS}(I_{Rain})) \tag{1}$$

其中 $F_{BS}(\cdot)$ 表示双线性下采样，以生成雨天图像样本 $I_{Rain,S}$。$\vartheta_{IDN}(\cdot)$ 表示IDN中的雨水评估函数。

雨水分布展示了退化的位置和程度，将其转化为退化自然是合理的，有助于准确的恢复背景。在将 $I^*_{B,S}$ 传入BRN进行背景重构之前，设计了一个多输入注意力模块（MAM），如图3所示，该模块通过Transformer层能够充分利用来自雨天图像 $I_{Rain}$ 的背景信息进行互补，并将其合并为嵌入表征 $I^*_{B,S}$。MAM的流程表示为：

$$f_{BT} = F_{SA}(I^*_{B,S}, I_{Rain}), \tag{2}$$

$$f_{MAM} = F_{HBF}(f_{BT}, F_B(I^*_{B,S})) \tag{3}$$

在公式（2）中，$F_{SA}(\cdot)$ 表示自注意力函数，包含了嵌入函数和点乘交互。$F_B(\cdot)$ 是生成初始表征 $I^*_{B,S}$ 的嵌入函数。$F_{HBF}(\cdot)$ 是指HFB中的融合功能。之后，BRN将 $f_{MAM}$ 作为背景的重构：

$$I^*_B = \vartheta_{BRN}(f_{MAM}) + F_{UP}(I^*_{B,S}) \tag{4}$$

其中 $\vartheta_{BRN}(\cdot)$ 表示BRN的超分辨率重建函数，$F_{UP}(\cdot)$ 表示双线性上采样。

与单独训练雨纹消除和背景重构不同，本文引入了联合约束来增强去雨模型与背景重构的兼容性，且能够从训练数据中自动进行学习。然后使用图像损失（Charbonnier损失函数（Hu等，2022；Jiang等，2022；Lai等，2017））和结构相似性（SSIM（Wang等，2004））损失对网络进行监督学习，同时实现图像和结构保真度的恢复。损失函数表示为：

$$L_{IDN} = \sqrt{(I^*_{B,S} - I_{B,S})^2 + \varepsilon^2} + \alpha \times SSIM(I^*_{B,S}, I_{B,S}) \tag{5}$$

$$L_{BRN} = \sqrt{(I^*_B - I_B)^2 + \varepsilon^2} + \alpha \times SSIM(I^*_B, I_B) \tag{6}$$

$$L = L_{IDN} + \lambda \times L_{BRN} \tag{7}$$

其中 $\alpha$ 和 $\lambda$ 用于平衡损失成分，分别设置为−0.15和1。惩罚系数 $\varepsilon$ 设置为1e-3。

## 2.2 交叉融合网络

众所周知，自注意力机制是Transformer的核心部分，它擅长学习长距离的语义依存关系和捕捉图像中的全局表示。与之相反，由于固有的局部连通性，CNNs更加擅长对局部关系进行建模。为此，本

文结合Transformer和CNNs的优势，将IDN和BRN的构建成深度双分支交叉融合网络。如图3所示，主干包括残差Transformer分支（RTB）和编码器-解码器分支（EDB）。RTB以一些可学习的内容（特征通道）作为输入，叠加多头注意力和前馈网络来编码全局结构。然而，获取长距离像素的相互关系是造成Transformer计算量巨大的罪魁祸首，使其无法应用于高分辨率图像，尤其是图像重构任务。受（Ali等，2021）启发，除了在样本空间上处理表征外，本文没有学习全局的空间相似性，而是应用SA计算跨通道的互协方差，以生成隐式编码全局上下文的注意力图，它具有线性复杂度而不是二次复杂度。

EDB旨在推理局部中丰富的纹理，受U-Net（Ronneberger等，2015）的启发，还使用U形框架构建了EDB。将前三个阶段构成编码器，其余三个阶段作为解码器。每个阶段采用类似的架构，包括采样层、残差通道注意块（Residual Channel Attention Blocks，RCABs）（Zhang等，2018）和交叉融合块。使用双线性采样和1×1卷积层来减少棋盘伪影和模型参数，而不是使用跨步或转置卷积来重新缩放特征的空间分辨率。为了促进不同阶段或尺度下的残差特征融合，设计了HFB以在空间和通道维度上聚合不同阶段的多个输入。HFB可以在重构过程中充分利用更多不同的功能。此外，为了进一步减少参数量，RTB和EDB设置了深度可分离卷积（Depth-wise Separable Convolutions，DSC）。对于RTB，将DSC集成到多头注意力中，以在计算特征协方差之前强调局部上下文，从而生成全局注意图。此外，将EDB构造成非对称U形结构，其中编码器设计了便携式的DSC，但解码器使用标准卷积。该方案可以节省整个网络约8%的参数。实验证明，在编码器中使用DSC的编码器优于在解码器使用。

### 2.3 多输入注意力模块

如图3，为将雨纹去除和背景重构联系起来，本文构建了一个带有Transformer的多输入注意力模块MAM，充分利用背景信息进行互补增强。不同于将系列图像块作为Transformer的输入，MAM将预测的雨水分布 $I^{*}_{R,S}$ ，子空间的去雨图像 $I^{*}_{B,S}$ 和雨天图像 $I_{Rain}$ 作为输入，首先学习嵌入表征 $(f^{*}_{B,S}, f^{*}_{R,S}, f_{Rain})$ 去丰富局部语义内容。$f^{*}_{R,S}$ 和 $f_{Rain}$ 查询（query，Q），键（key，K）和值（value，V）的映射。这里不对大小为 $R^{HW\times HW}$ 的空间注意图进行学习，而是重新定义Q和K的映射大小，并通过 $f^{*}_{R,S}$ 和 $f_{Rain}$ 之间的点乘，生成交叉的协方差转置注意力图 $M\in R^{C\times C}$ 。

如图5所示，注意力图引导网络从 $I_{Rain}$ 的嵌入表征 $f_{Rain}$ 中挖掘背景纹理信息 $f_{BT}$ 。SA的处理流程表示为：

$$F_{\mathrm{SA}}=(\mathrm{Soft\,max}(F_{\mathrm{K}}(I^{*}_{R,S})\circ F_{\mathrm{Q}}(I_{Rain})))\circ F_{\mathrm{V}}(I_{Rain}), \qquad (8)$$

其中 $F_{\mathrm{K}}(\cdot)$ ，$F_{\mathrm{Q}}(\cdot)$ 和 $F_{\mathrm{V}}(\cdot)$ 是进行映射的嵌入函数，$\circ$ 是点乘操作。之后在交叉混合模块中，将提取的互补信息和 $I^{*}_{B,S}$ 的嵌入表征结合去丰富背景表征。

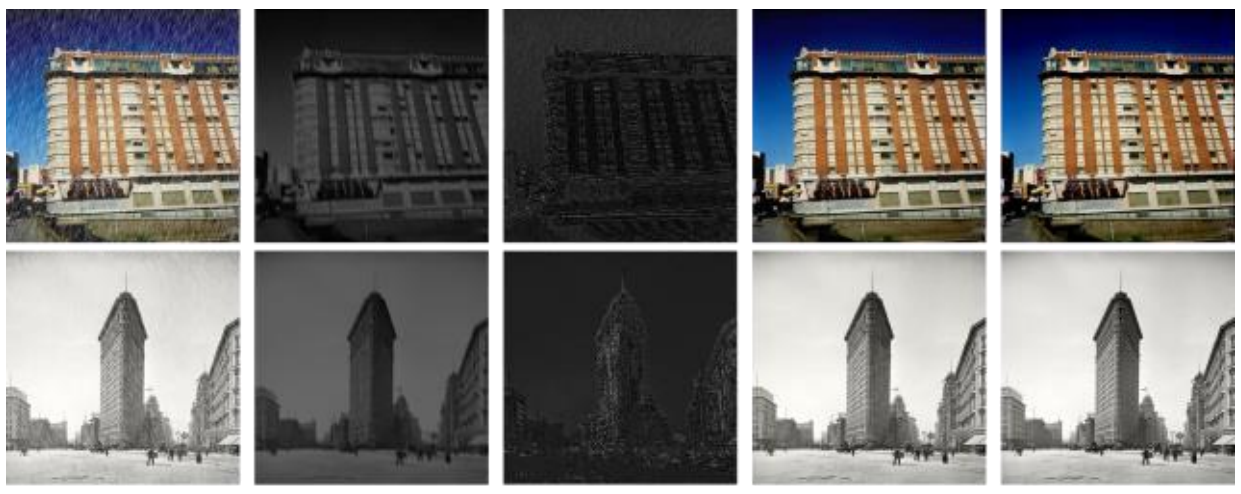

图 5 MAM 的可视化

Fig.5 Visualization of MAM

### 2.4 交叉融合模块

考虑到残差块和编码阶段之间的特征冗余和知识差异，本文引入了一种新的交叉融合块 HFB，其中早期阶段的低层次背景特征有助于巩固后期阶段的高层次特征。具体来说，将深度可分离的卷积和通道注意层纳入 HFB，以便在空间和通道维度上辨别性地聚合多尺度特征。与基于像素级叠加或卷积融合相比，提出的 HFB 更加灵活和有效。

## 3 实验结果

为了验证本文提出的 ELF，在合成和真实的雨天数据集上进行了广泛的实验，并将 ELF 与几种主流的图像去雨方法进行了比较。这些方法主要包含了 MPRNet（Zamir 等，2021）、SWAL（Huang 等，2021）、RCDNet（Wang 等，2020）、DRDNet（Deng等，2020）、MSPFN（Jiang 等，2020）、IADN（Jiang等，2020）、PreNet（Ren 等，2019）、UMRL（Yasarla等，2019）、DIDMDN（Zhang 等，2018）、RESCAN（Li 等，2018）和 DDC（Li 等，2019）。本节使用五种常用的评估指标进行评测，例如峰值信噪比（Signal to Noise Ratio，PSNR）、结构相似性（Structural Similarity，SSIM）、特征相似性（Feature Similarity，FSIM）、自然度图像质量评估器（Naturalness Image Quality Evaluator，NIQE）（Mittal等，2012）和基于空间熵的质量（Spatial-Spectral

Entropy-based Quality，SSEQ）（Liu 等，2014）。

### 3.1 实验细节

#### 3.1.1 数据收集

由于所有比较方法的训练样本存在差异，根据（Jiang 等，2020），本节使用（Fu 等，2017；Zhang 等，2019）中的 13700 个干净的背景/雨天图像对，用其公开发布的代码训练所有比较方法，并通过调整优化参数以进行公平比较。在测试阶段，选取了四个合成基准（Test100（Zhang 等，2019）、Test1200（Zhang 等，2018）、R100H 和 R100L（Yang 等，2017））和三个真实数据集（Rain in Driving（RID）、Rain in SurveilLance（RIS）（Li 等，2019）和 Real127（Zhang 等，2018））进行评估。

#### 3.1.2 实验设置

在本文的基线中，RTB的Transformer模块数量设为10，根据经验，对于EDB中的每个阶段，RCAB设置为1，滤波器数量为48。为了方便训练，将训练图像裁剪为固定大小为256×256像素的块，以获得训练样本。使用学习率为2e-4的Adam优化器，每65个训练轮数的衰减率为0.8，直到600轮。批量大小为12，在单个Titan Xp GPU上训练ELF模型500轮次。

### 3.2 消融研究

#### 3.2.1 基本组件的验证

为了验证网络中各个组件对最终去雨性能的贡献，本节进行了相应的消融研究，包括自注意力（SA）、深度可分离卷积（DSC）、超分辨率重构（SR）、交叉融合模块（HFB）和多输入注意力模块（MAM）。为了简单起见，将最终模型表示为 ELF，并通过删除上述所有组件来表示基线模型 *w/o*。在 Test1200 数据集上的去雨性能和推理效率方面的定量结果如表 1 所示，实验表明完整的去雨模型 ELF 比其不完整的版本有着显著的改进。与 *w/o MAM*（从 ELF 中删除 MAM 模块）模型相比，ELF 实现了 1.92dB 的性能增幅，主要是因为 MAM 中的关联学习可以帮助网络充分利用雨天输入的背景信息和预先预测的雨水分布。此外，将图像去雨任务分解为低维空间的雨纹去除和纹理重建在效率（推理时间和计算成本分别上升了 19.8%和 67.6%）和重构质量（参考 ELF 和 ELF*的结果，ELF*在原始分辨率空间上完成去雨和纹理重构）上具有相当大的优势。使用深度可分离卷积可以在参数大致相同的情况下增加通过深度，从而提高表示能力（参考 ELF 和 *w/o DSC* 模型的结果）。与用标准 RCABs 替换 RTB 中的 Transformer 块的 *w/o SA* 模型相比，ELF 在可接受的计算成本下提升了 0.45dB。

本节进行了双分支交叉融合框架的消融实验，其中涉及一个残差 Transformer 分支（RTB）和一个 U 型编码器-解码器分支（EDB）。基于 ELF，依次去除这两个分支，设计两个对比模型（*w/o RTB* 和 *w/o EDB*），定量结果如表 1 所示。去除 RTB 可能会大大削弱对空间结构的表示能力，导致性能大幅下降（参考 ELF 和 *w/o RTB* 模型的结果，PSNR 下降 2.09dB）。此外，EDB 允许网络聚合多尺度的纹理特征，这对于丰富局部纹理的表征至关重要。

**表1 在Test1200数据集上消融实验。✓表示加入该模块，×表示不加入该模块。**

**Table 1 Ablation study on Test1200 dataset. ✓ and × denote adding and not adding the modules, respectively.**

| Model | SA | SR | DSC | HFB | MAM | SSIM | PSNR(dB) | SSIM | Par. (M) | Time | Gflops |
|---|---|---|---|---|---|---|---|---|---|---|---|
| Rain Image | - | - | - | - | - | - | 22.16 | 0.732 | - | - | - |
| *w/o* SA | × | ✓ | ✓ | ✓ | ✓ | ✓ | 32.78 | 0.919 | 1.536 | 50.24 | 50.24 |
| *w/o* DSC | ✓ | ✓ | × | ✓ | ✓ | ✓ | 32.73 | 0.918 | 1.518 | 0.121 | 62.37 |
| *w/o* HFB | ✓ | ✓ | ✓ | × | ✓ | ✓ | 32.56 | 0.917 | 1.539 | 0.102 | 69.41 |
| *w/o* MAM | ✓ | ✓ | ✓ | ✓ | × | ✓ | 31.46 | 0.906 | 1.516 | 0.121 | 64.28 |
| *w/o* SSIM | ✓ | ✓ | ✓ | ✓ | ✓ | × | 33.17 | 0.919 | 1.532 | 0.125 | 66.39 |
| *w/o* all | × | ✓ | × | × | × | × | 29.05 | 0.861 | 1.538 | 0.046 | 61.24 |
| ELF* | ✓ | × | ✓ | ✓ | × | ✓ | 32.97 | 0.921 | 1.519 | 0.156 | 214.67 |
| ELF | ✓ | ✓ | ✓ | ✓ | ✓ | ✓ | 33.38 | 0.925 | 1.532 | 0.125 | 66.39 |

| Model | RTB | EDB | PSNR | SSIM | Par. | Time | Gflops |
|---|---|---|---|---|---|---|---|
| *w/o* RTB | × | ✓ | 31.29 | 0.910 | 1.536 | 0.041 | 26.48 |
| *w/o* EDB | ✓ | × | 33.04 | 0.922 | 1.534 | 0.154 | 134.27 |
| ELF | ✓ | ✓ | 33.38 | 0.925 | 1.532 | 0.125 | 66.39 |

表2 在四个数据集上PSNR、SSIM和FSIM的比较结果。

Table2 Comparison results of average PSNR, SSIM, and FSIM on four datasets.

| Methods | RESCAN | UMRL | PreNet | IADN | MSPFN | DRDNet | PCNet | MPRNet | SWAL | ELF-LW | ELF |
|---|---|---|---|---|---|---|---|---|---|---|---|
| Datasets | | | | | | **Test100/Test1200** | | | | | |
| PSNR (dB) | 25.00/30.51 | 54.41/30.55 | 24.81/31.36 | 26.71/32.29 | 27.50/32.39 | 28.06/26.73 | 26.17/32.03 | 30.27/32.91 | 28.47/30.40 | 29.41/32.55 | **30.45/33.38** |
| SSIM | 0.835/0.882 | 0.829/0.910 | 0.851/0.911 | 0.865/0.916 | 0.876/0.916 | 0.874/0.824 | 0.871/0.913 | 0.897/0.916 | 0.889/0.892 | 0.894/0.912 | **0.909/0.925** |
| FSIM | 0.909/0.944 | 0.910/0.955 | 0.916/0.955 | 0.924/0.958 | 0.928/0.960 | 0.925/0.920 | 0.924/0.956 | 0.939/0.960 | 0.936/0.950 | 0.937/0.960 | **0.945/0.964** |
| Datasets | | | | | | **R100H/R100L** | | | | | |
| PSNR (dB) | 26.36/29.80 | 26.01/29.18 | 26.77/32.44 | 27.86/32.53 | 28.66/32.40 | 21.21/29.24 | 28.45/34.42 | 30.41/36.40 | 29.30/34.60 | 28.83/34.61 | **30.48/36.67** |
| SSIM | 0.786/0.881 | 0.832/0.923 | 0.858/0.950 | 0.835/0.934 | 0.860/0.933 | 0.668/0.883 | 0.871/0.952 | 0.889/0.965 | 0.887/0.958 | 0.876/0.958 | **0.896/0.968** |
| FSIM | 0.864/0.919 | 0.876/0.940 | 0.890/0.956 | 0.875/0.942 | 0.890/0.943 | 0.797/0.903 | 0.897/0.959 | 0.910/0.969 | 0.908/0.963 | 0.901/0.962 | **0.915/0.972** |
| Avg-PSNR↑ | 27.91 | 27.53 | 28.84 | 29.84 | 30.23 | 26.31 | 30.27 | 32.49 | 30.69 | 31.35 | **32.74** |
| Par.(M)↓ | 0.150 | 0.984 | 0.169 | 0.980 | 13.350 | 5.230 | 0.655 | 3.637 | 156.540 | **0.566** | 1.532 |
| Time (S)↓ | 0.546 | 0.112 | 0.163 | 0.132 | 0.507 | 1.426 | **0.062** | 0.207 | 0.116 | 0.072 | 0.125 |
| GFlops(G)↓ | 129.28 | 65.74 | 265.76 | 80.99 | 708.39 | 896.43 | 28.21 | 565.81 | 614.35 | **21.53** | 66.39 |

注：**加粗数字**表示最优结果。

## 3.3 与 SOTA 的比较

### 3.3.1 合成数据

表 2 提供了在 Test1200、Test100、100H 和 R100L 数据集的定量结果，以及推理时间、模型参数和计算成本。据观察，大多数模型在小雨的情况下一致地获得了优异性能，而 ELF 和 MPRNet 在大雨条件下仍然表现良好，特别在 PSNR 上显示出更大的优势。ELF 模型在所有指标上都取得最优，平均超过了基于 CNN 的 SOTA（MPRNet）0.25 dB，且仅占其计算成本和参数的 11.7%和 42.1%。同时，轻量模型 ELF-LW 仍然具有竞争力，在四个数据集上的 PSNR 分数排名第三，平均比实时图像去雨的方法 PCNet（Jiang 等，2021）高出 1.08dB，并具有更少的参数（节省 13.6%）和计算成本（节省 23.7%）。图 6 提供了结果可视化，PreNet、MSPFN 和 RCDNet 等高精度的方法，可以有效消除雨水层，从而提高可见度。但由于大量的伪影和不自然的颜色外观，尤其是在大雨条件下，它们未能在视觉上产生好的效果。DRDNet 专注于细节的重构，但推理过程耗时长、内存大。MPRNet 往往会产生过度平滑的结果。ELF 除了重构出更干净和更可靠的图像纹理外，产生的结果也具有更好的对比度以及更少的颜色失真（参考“老虎”和“马”的场景）。此外，可以推出重构质量的改善可能得益于提出的 Transformer 和 CNN 的混合表示框架以及用于雨纹去除和背景重构的关联学习方案。这些策略方法被集成到一个统一的框架中，使得网络能够充分利用各自的学习优点进行图像去雨，同时保证模型的推理效率。

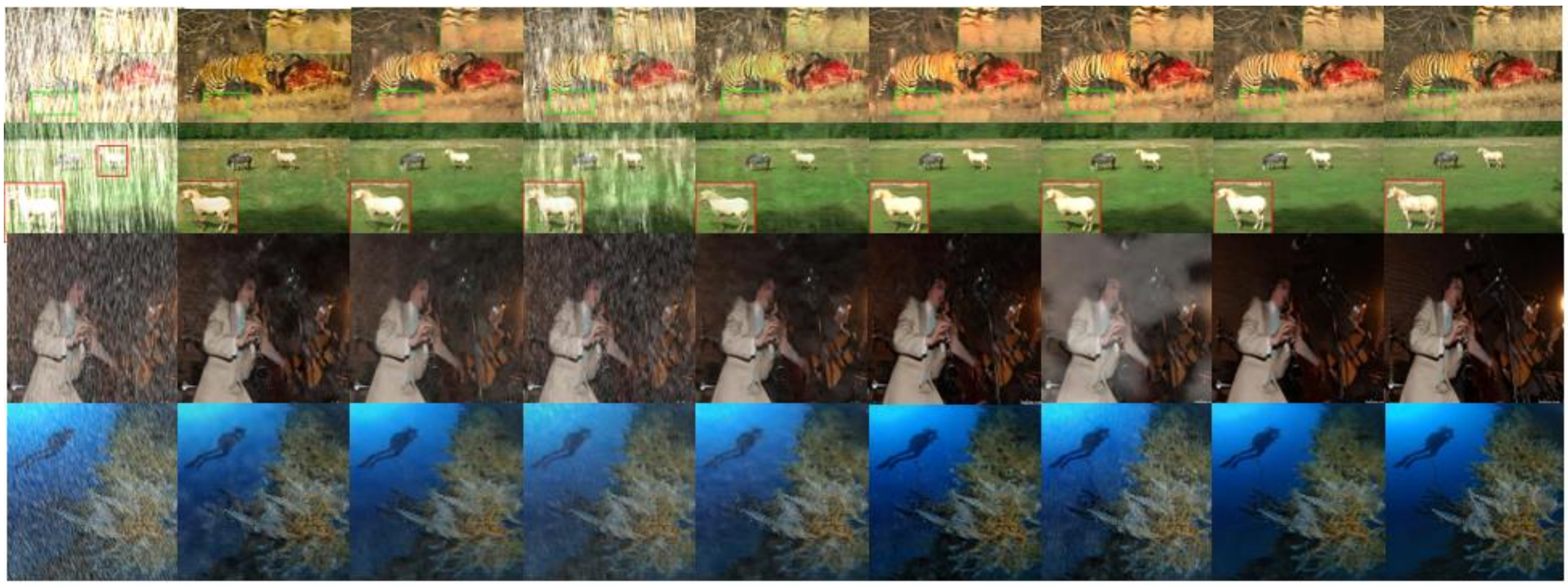

输入 PreNet MSPFN DRDNet RCDNet MPRNet SWAL ELF(Ours) 真实标签

图 6 七种图像去雨方法的可视化结果对比

Fig.6 Visualization comparison of the results of 7 image deraining methods.

3.3.2 真实场景数据

本节进一步地在三个真实场景的数据集上进行了实验：Real127、Rain in Driving（RID）和 Rain in Surveillance（RIS）。表 3 列出了 NIQE 和 SSEQ 的定量结果，其中 NIQE 和 SSEQ 分数越小，表示感知的质量越好，内容越清晰。ELF 同样具有很强的竞争力，在 RID 数据集上的平均分数值最低，NIQE 和 SSEQ 的平均分数在 Real127 和 RIS 数据集上是最好的。图 7 直观地展示了得出的结果，可以看出 ELF 产生的无雨图像中内容更干净、更可信，而其他的方法未能很好的去除雨痕。这些实验表明了 ELF 模型不仅能够很好的消除雨水扰动，同时还能保留纹理细节和图像自然度。

**表3 三个真实数据集上十种图像去雨方法的NIQE/SSEQ平均分数对比。**

**Table3 Comparison of average NIQE/SSEQ scores with ten deraining methods on three real-world datasets.**

| Datasets | DIDMDN | RESCAN | DDC | LPNet | UMRL | PreNet | IADN | MSPFN | DRDNet | MPRNet | ELF |
|---|---|---|---|---|---|---|---|---|---|---|---|
| Real127 (127) | 3.929/ | 3.852/ | 4.022/ | 3.989/ | 3.984/ | 3.835/ | 3.769/ | 3.816/ | 4.208/ | 3.965/ | **3.735/** |
| | 32.42 | 30.09 | 29.33 | 29.62 | 29.48 | 29.61 | 29.12 | **29.05** | 30.34 | 30.05 | 29.16 |
| RID (2495) | 5.693/ | 6.641/ | 6.247/ | 6.783/ | 6.757/ | 7.007/ | 6.035/ | 6.518/ | 5.715/ | 6.452/ | **4.318/** |
| | 41.71 | 40.62 | 40.25 | 42.06 | 41.04 | 43.04 | 40.72 | 40.47 | 39.98 | 40.16 | **37.89** |
| RIS (2348) | 5.751/ | 6.485/ | 5.826/ | 6.396/ | **5.615/** | 6.722/4 | 5.909/ | 6.135/ | 6.269/ | 6.610/ | 5.835/ |
| | 46.63 | 50.89 | 47.80 | 53.09 | 43.45 | 8.22 | 42.95 | 43.47 | 45.34 | 48.78 | **42.16** |

注：**加粗数字**表示最优结果。

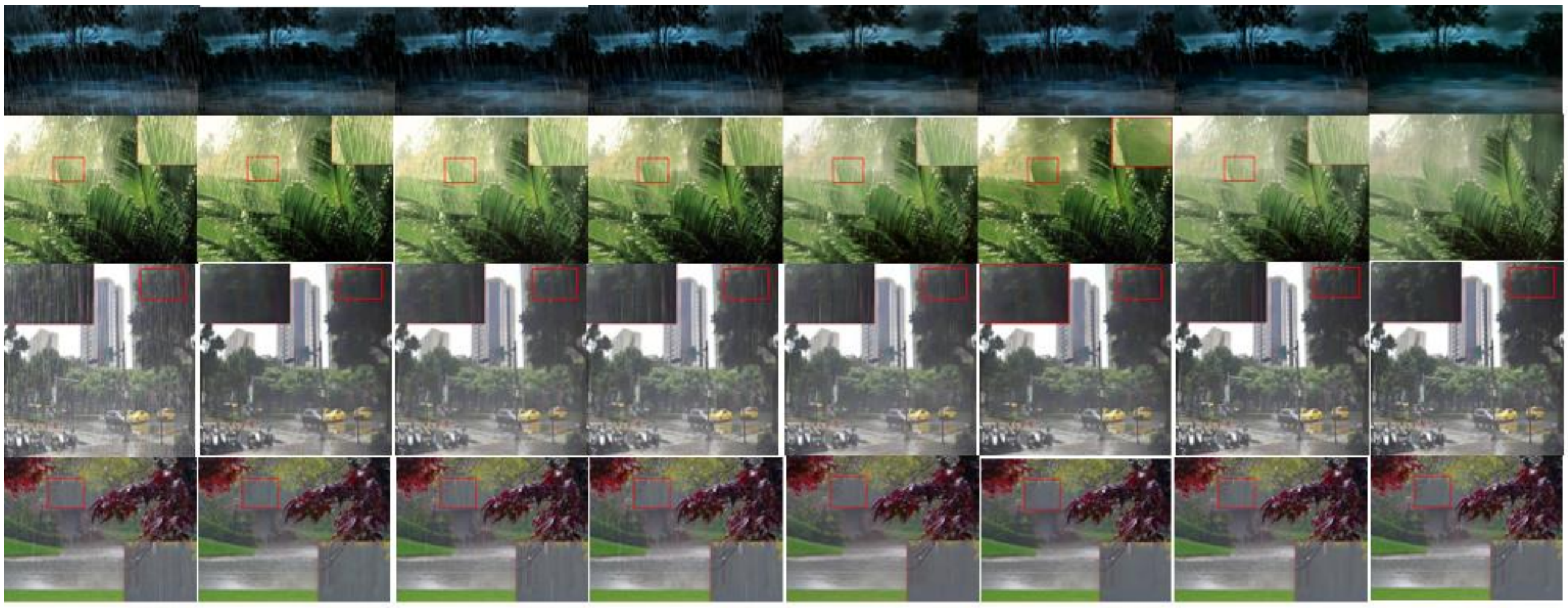

图 7 八种方法在五个真实场景中去雨的结果对比，包含了雨雾效应，大雨和小雨

Fig.7 Visual comparison of derained images obtained by eight methods on five real-world scenarios, coverring rain veiling, heavy rain and light rain.

### 3.4 对下游视觉任务的影响

在雨天条件下消除雨纹的退化影响，同时保留可靠的纹理细节对于目标检测来说至关重要。这就促使本文研究去雨对目标检测算法中检测精度的影响。为此，将 ELF 和几个有代表性的去雨方法直接应用在一些雨天图像并生成对应的无雨图像，然后使用公开的 YOLOv3（Redmon 等，2018）预训练模型进行检测。表 4 表明了 ELF 在 COCO350 和 BDD350 数据集（Jiang 等，2020）上的 PSNR 分数最高，与其他去雨方法相比，ELF 生成的无雨结果具有更好的目标检测性能。图 8 中两个样本的比较表明，ELF 去雨图像在图像质量和检测精度方面有着显著的优势。去雨和下游检测任务的显著性能归因于雨纹消除和细节重构任务之间的关联学习。

### 3.5 对其他图像恢复任务的通用性

一些图像恢复任务如水下图像增强，低光照图像增强等，具有和图像去雨相似的退化干扰因素，因此，为进一步探索提出的ELF的通用性与稳定性，本节在水下图像增强和低光照图像增强任务上开展了简单的研究。

3.5.1 水下图像增强

根据（Li等，2021），使用2050对生成的水下图像来训练ELF。其中，800对图像选自UIEB（Li等，2019）数据集，1250对图像选自（Li等，2020）中提出的数据集S1000。分别在真实场景数据集R90（Li等，2019）和合成数据集S1000上进行了实验，并和7个主流的水下图像增强方法进行了对比。

**表4 在COCO350/BDD350数据集上联合图像去雨和目标检测的结果比较。**

**Table4 Comparison results of joint image deraining and object detection on COCO350/BDD350.**

| Methods | Rain input | RESCAN | PreNet | IADN | MSPFN | MPRNet | ELF-LW | ELF |
|---|---|---|---|---|---|---|---|---|
| Deraining; Dataset: **COCO350/BDD350**; Image Size: **640×480/1280×720** | | | | | | | | |
| PSNR (dB) | 14.79/14.13 | 17.04/16.71 | 17.53/16.90 | 18.18/17.91 | 18.23/17.85 | 17.99/16.83 | 18.43/18.09 | **18.93**/**18.49** |
| SSIM | 0.648/0.470 | 0.745/0.646 | 0.765/0.652 | 0.790/0.719 | 0.782/**0.761** | 0.769/0.622 | 0.800/0.714 | **0.818**/**0.761** |
| Ave.inf.time(s) | –/– | 0.546/1.532 | 0.227/0.764 | 0.135/0.412 | 0.584/1.246 | 0.181/0.296 | **0.076**/**0.160** | 0.128/0.263 |
| Object Detection; Algorithm: **YOLOv3**; Dataset: **COCO350/BDD350**; Threshold: 0.6 | | | | | | | | |
| Precision (%) | 23.03/36.86 | 28.74/40.33 | 31.31/38.66 | 32.92/40.28 | 32.56/41.04 | 31.31/40.49 | 33.31/40.88 | **33.85/41.84** |
| Recall (%) | 29.60/42.80 | 35.61/47.79 | 37.92/48.59 | 39.83/50.25 | 39.31/50.40 | 38.98/48.77 | 40.43/51.55 | **40.43/52.60** |
| IoU (%) | 55.50/59.85 | 59.81/61.98 | 60.75/61.08 | **61.96**/62.27 | 61.69/**62.42** | 61.14/61.99 | 61.14/61.99 | 61.14/61.99 |

注：**加粗数字**表示最优结果。

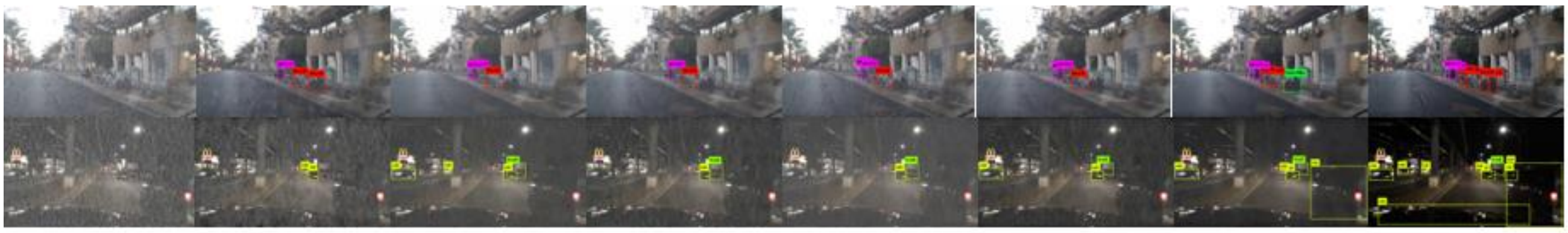

图 8 在 BDD350 数据集上联合图像去雨和目标检测的可视化比较

Fig.8 Visual comparison of joint image deraining and object detection on BDD350 dataset.

**表5 在S1000和R90数据集上比较七种水下图像增强方法的PSNR (dB)/MSE(×10³)平均分数。**

**Table5 Comparison of average PSNR (dB)/MSE (×10³) scores with seven underwater image enhancement methods on S1000 and R90 datasets.**

| Methods | input | Guo 等，2020 | UcycleGAN | Water-Net | UWCNN_retrain | Unet-U | Unet-RMT | Ucolor | ELF |
|---|---|---|---|---|---|---|---|---|---|
| S1000 | 12.96 | 15.78 | 14.73 | 15.47 | 15.87 | 19.14 | 17.93 | <u>23.05</u> | **27.20** |
| | /4.60 | /2.57 | /3.13 | /3.26 | /2.74 | /1.22 | /1.43 | /<u>0.50</u> | /**0.32** |
| R90 | 16.11 | 18.05 | 16.61 | 19.81 | 16.69 | 18.14 | 16.89 | <u>20.63</u> | **24.57** |
| | /2.03 | /1.18 | /1.65 | /1.02 | /1.71 | /1.32 | /1.71 | /<u>0.77</u> | /**0.44** |

注：**加粗数字**表示最优结果，<u>下划线数字</u>表示次优结果。

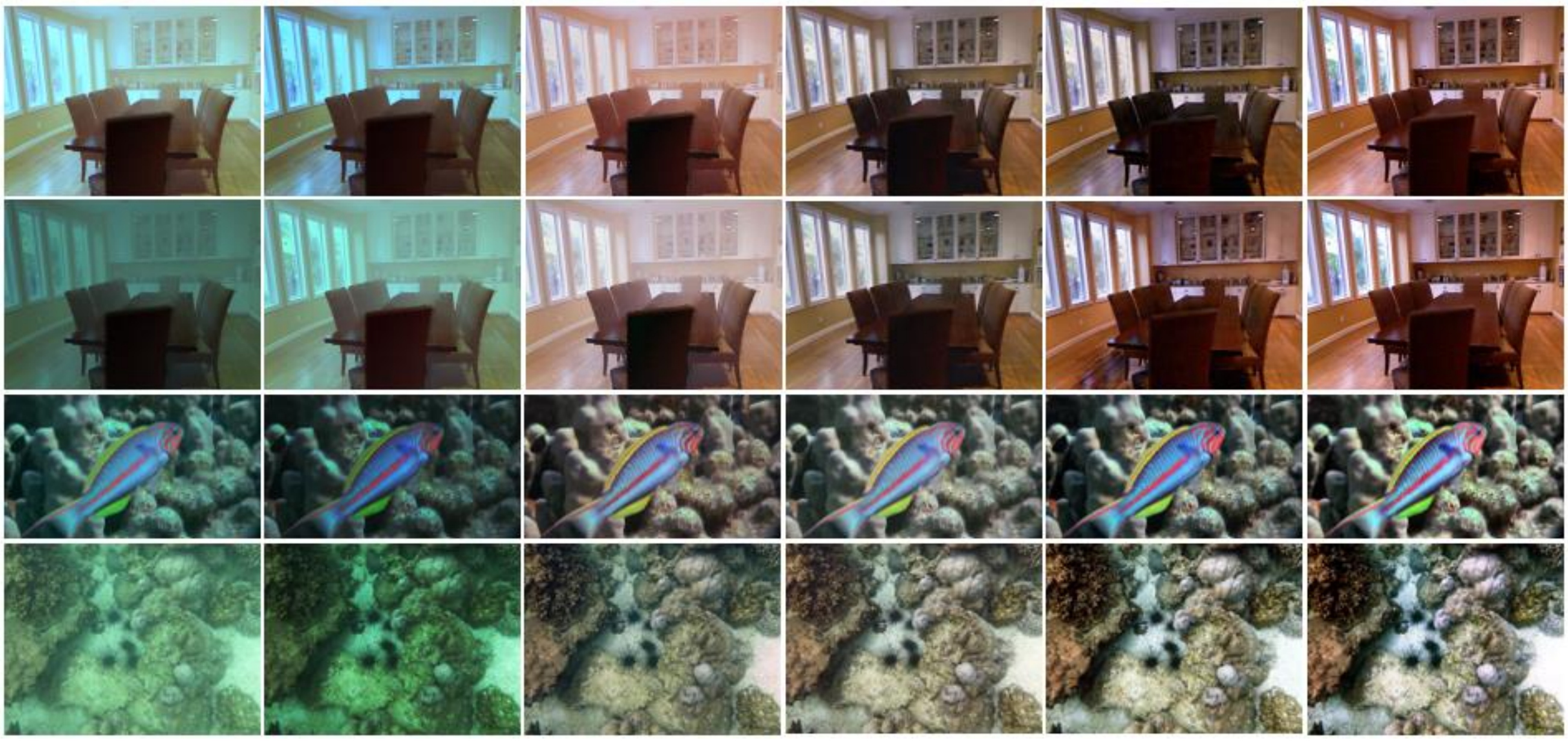

图 9 五种方法在 S1000（前两行）和 R90（后两行）数据集上增强后的结果比较

Fig.9 Visual comparison of enhanced images obtained by five methods on S1000 and R90.

表 5 列出了 PSNR 和 MSE（Mean Squared Error）的定量结果，其中 PSNR 分数越大，MSE 分数越小表明图像的质量越好。可以看到，ELF 在 R90 和 S1000 数据集上都取得了最好的结果，且平均 PSNR 分数比 UColor（Li 等，2021）方法分别高出 4.15dB 和 3.94dB。图 9 直观地展示了得出的结果，可以看到，提出的方法在有效矫正水下图像的对比度和光照失真，同时可以恢复出了更真实的细节结构，而其他对比方法，要么没有消除水下异常色调，要么恢复出的图像模糊缺少细节信息。这些实验表明了 ELF 模型在水下图像增强任务上的有效性和优势。

### 3.5.2 低光照图像增强

本节在低光照图像增强领域最常用的基准数据集之一 LOL（Li 等，2019）数据集上进行实验，评估 ELF 在该任务上的性能。使用 LOL 训练集中 485 对低光-正常光的图像对训练 ELF，并在测试集上进行测试。除了 PSNR 和 SSIM 评估指标外，还使用了 LPIPS（Learned Perceptual Image Patch Similarity）（Li 等，2020）指标，越低的 LPIPS 分数表示感知质量越好。表 6 给出了和 9 个主流的低光照增强方法的定量对比结果，ELF 取得了最高的 PSNR 分数，在 SSIM 和 LPIPS 指标上也十分接近当下的最优方法。其中，LLFolw 采用额外的条件编码器提取光照不变的颜色图作为先验分布的均值，并利用基于低光照图像/特征为条件的负对数似然损失，这有助于表征图像结构和上下文内容，保证在图像流形中具有和真值相近的颜色分布。因此，LLFolw 可以获得更好的 MAE 和 LPIPS 分数。相比之下，本文提出的 ELF 不需要任何颜色先验，但因具有精细的关联学习方案，和逐像素、结构一致性约束，这对本文方法获得更高的 PSNR 得分贡献更大。

为了进一步显示 ELF 的有效性，图 10 展示了直观的视觉结果。可以看到，部分方法恢复出的图像存在较大的噪声和伪影，如 EnlightenGAN（Jiang 等，2021）和 KinD++（Zhang 等，2021）；一些方法要么增强后的亮度不足，要么出现了过曝光的情况；相比之下，ELF 在合理增强图像亮度的同时，受噪声和色偏的影响较小，且恢复出了更接近原图的结构信息。这些实验表明了 ELF 模型在低光图像增强任务上的鲁棒性，也验证了提出的退化消除和背景恢复关联学习方案的有效性。

**表6 在LOL数据集上比较九种低光照图像增强方法的PSNR (dB)、SSIM、LPIPS和MAE(%)。**

**Table6 Comparison of average PSNR (dB), SSIM, LPIPS and MAE (%) scores with nine low light image enhancement methods on LOL datasets.**

| Methods | LIME | RetinexNet | Zero-DCE | EnlightenGAN | KinD | KinD++ | DCC-Net | LLFlow | ELF |
|---|---|---|---|---|---|---|---|---|---|
| PSNR (dB) | 16.67 | 16.77 | 16.80 | 17.48 | 20.87 | 21.30 | 22.72 | <u>25.19</u> | **26.28** |
| SSIM | 0.56 | 0.56 | 0.54 | 0.65 | 0.80 | 0.82 | 0.81 | **0.93** | <u>0.92</u> |
| LPIPS | 0.35 | 0.47 | 0.33 | 0.32 | 0.17 | 0.16 | 0.14 | **0.11** | <u>0.10</u> |
| MAE | 12.00 | 12.56 | 13.70 | 13.71 | 9.82 | 8.83 | 8.72 | **5.85** | <u>8.07</u> |

注：**加粗数字**表示最优结果，<u>下划线数字</u>表示次优结果。

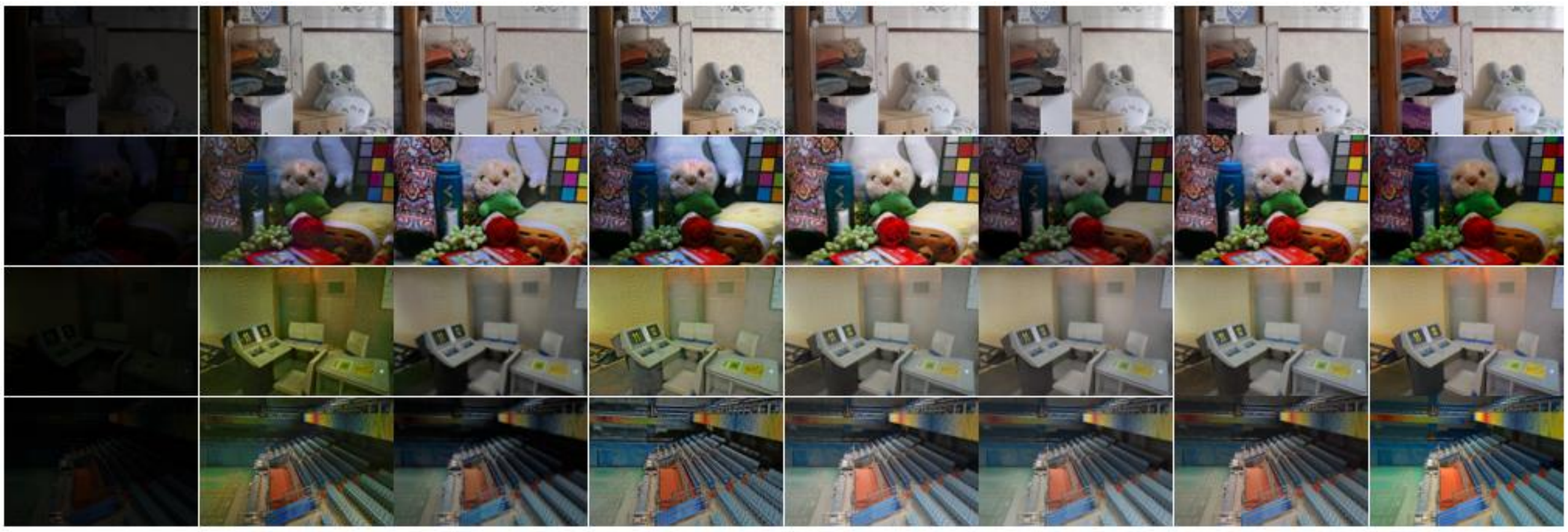

输入　EnlightenedGAN　KinD　KinD++　DDCNet　LLFLOW　Ours　真实标签

图 10 七种方法在 LOL 数据集上增强后的结果比较

Fig.10 Visual comparison of enhanced images obtained by seven methods on LOL dataset

## 4 结论

基于降质分布揭示了图像退化位置和程度的观察，本文引入退化先验来帮助精确的背景恢复，并据此提出了高效高质的部分-整体图像扰动去除和背景修复方案，即ELF。为了在提高模型紧凑型的同时实现关联学习，本文提出同时利用Transformer和CNNs的优势，构建一个精心设计的多输入注意力模块（Multi-input Attention Module, MAM）来实现扰动去除和背景修复的关联学习。在图像去雨、水下图像增强、低光图像增强和联合目标检测任务上的大量实验结果表明，本文提出的ELF模型远优于现有的主流图像增强模型。

尽管本文提出的方法在图像去雨、水下图像增强、低光图像增强，以及联合目标检测任务上展示了令人印象深刻的效果，但因缺少对不同降质的特性和共性特征的特定表达，在应对具有多种复杂天气条件的真实场景时可能失效。同时，该方法仍然需求大量的高质量成对训练数据，极大限制了在新场景和新任务上的推广，并且和真实环境降质存在极大的域差异。为了解决上述问题，未来作者团队拟引入视觉大模型作为特征、语义表达先验，在隐式空间实现不同降质环境下场景本质信息的表征，消除场景和降质差异；进一步引入基于提示的文本语言大模型，实习实时可交互的场景内容修复、理解和分析。

## 参考文献（References）

## 作者简介

江奎，男，副教授，主要研究方向为计算机视觉与图像增强。E-mail：kuijiang_1994@163.com

黄文心，女，讲师，主要研究方向为计算机视觉与多模态检索。E-mail：wenxinhuang_wh@163.com

贾雪梅，女，博士研究生，主要研究方向为图像、视频处理与可信计算。E-mail：jiaxuemeiL@163.com

王文兵，男，硕士研究生，主要研究方向为计算机视觉与图像增强。E-mail：wenbing.wang@rokid.com

王正，男，教授，主要研究方向为多模态信号处理与可信计算。E-mail：wangzwhu@whu.edu.cn

江俊君，男，教授，主要研究方向为计算机视觉与多媒体信号处理。E-mail：jiangjunjun@hit.edu.cn